%% file: acl2018.tex
\documentclass[11pt,a4paper]{article}
\usepackage[hyperref]{acl2018}
\usepackage{times}
\usepackage{latexsym}

\usepackage{url}

\usepackage{amsmath}
\usepackage{amssymb}
\usepackage{graphicx}
\newcommand{\cev}[1]{\reflectbox{\ensuremath{\vec{\reflectbox{\ensuremath{#1}}}}}}
\usepackage{framed}
\usepackage{subfigure}
\usepackage{booktabs}
\usepackage{empheq}
\usepackage{array}
\usepackage{color}
\usepackage{xcolor}
\usepackage{enumitem}
\usepackage{multirow}
\usepackage{mathtools}

\DeclareMathOperator*{\argmax}{arg\,max}

\DeclarePairedDelimiterX{\infdivx}[2]{(}{)}{%
	#1\;\delimsize\|\;#2%
}

\newcommand{\dKL}{D_{KL}\infdivx}

\newcommand{\ourModelName}{\textsc{NeuSum}}
\newcommand{\significant}{$ {}^{\text{\textbf{-}}} $}
\newcommand{\otherpaper}{$ {}^{\ddag} $}
\newcommand{\cnndm}{\textit{CNN/Daily Mail}}

\aclfinalcopy 

\title{Neural Document Summarization by Jointly \\Learning to Score and Select Sentences}

\author{Qingyu Zhou$^\dag$\thanks{\; Contribution during internship at Microsoft Research.},\, Nan Yang$^\ddag$,  Furu Wei$^\ddag$,  Shaohan Huang$^\ddag$,  Ming Zhou$^\ddag$,  Tiejun Zhao$^\dag$ \\
	$^\dag$Harbin Institute of Technology, Harbin, China \\
	$^\ddag$Microsoft Research, Beijing, China \\
	{\tt \{qyzhou,tjzhao\}@hit.edu.cn} \\ {\tt \{nanya,fuwei,shaohanh,mingzhou\}@microsoft.com}
}

\date{}

\begin{document}
\maketitle
\begin{abstract}
Sentence scoring and sentence selection are two main steps in extractive document summarization systems.
However, previous works treat them as two separated subtasks.
In this paper, we present a novel end-to-end neural network framework for extractive document summarization by jointly learning to score and select sentences.
It first reads the document sentences with a hierarchical encoder to obtain the representation of sentences.
Then it builds the output summary by extracting sentences one by one.
Different from previous methods, our approach integrates the selection strategy into the scoring model, which directly predicts the relative importance given previously selected sentences.
Experiments on the CNN/Daily Mail dataset show that the proposed framework significantly outperforms the state-of-the-art extractive summarization models.
\end{abstract}

\input{body/intro}

\input{body/related}

\input{body/problem}

\input{body/model}

\input{body/exp}
\input{body/diss}

\input{body/conclusion}

\input{body/ack}


\bibliography{acl2018}
\bibliographystyle{acl_natbib}


\end{document}

%% file: body/intro.tex
\section{Introduction}

Traditional approaches to automatic text summarization focus on identifying important content, usually at sentence level \cite{nenkova2011automatic}.
With the identified important sentences, a summarization system can extract them to form an output summary.
In recent years, \textit{extractive methods} for summarization have proven effective in many systems \cite{carbonell1998use,mihalcea2004textrank,mcdonald2007study,cao2015ranking}.
In previous works that use extractive methods, text summarization is decomposed into two subtasks, i.e., sentence scoring and sentence selection.

\textit{Sentence scoring} aims to assign an importance score to each sentence, and has been broadly studied in many previous works.
Feature-based methods are popular and have proven effective, such as 
word probability, TF*IDF weights, sentence position and sentence length features \cite{luhn1958automatic,hovy1998automated,Ren:2017:LCS:3077136.3080792}.
Graph-based methods such as TextRank \cite{mihalcea2004textrank} and LexRank  \cite{erkan2004lexrank} measure sentence importance using weighted-graphs.
In recent years, neural network has also been applied to sentence modeling and scoring \cite{cao2015ranking,Ren:2017:LCS:3077136.3080792}.

For the second step, \textit{sentence selection} adopts a particular strategy to choose content sentence by sentence.
Maximal Marginal Relevance  \cite{carbonell1998use} based methods select the sentence that has the maximal score and is minimally redundant with sentences already included in the summary.
Integer Linear Programming based methods \cite{mcdonald2007study} treat sentence selection as an optimization problem under some constraints such as summary length.
Submodular functions \cite{lin2011class} have also been applied to solving the optimization problem of finding the optimal subset of sentences in a document.
\citet{ren2016redundancy} train two neural networks with handcrafted features. One is used to rank  sentences, and the other one is used to model  redundancy during sentence selection.

In this paper, we present a neural extractive document summarization (\ourModelName{}) framework which jointly learns to score and select sentences.
Different from previous methods that treat sentence scoring and sentence selection as two tasks, our method integrates the two steps into one end-to-end trainable model.
Specifically, \ourModelName{} is a neural network model without any handcrafted features that learns to identify the relative importance of sentences.
The relative importance is measured as the gain over previously selected sentences.
Therefore, each time the proposed model selects one sentence, it scores the sentences considering both  sentence saliency and previously selected sentences.
Through the joint learning process, the model learns to predict the relative gain given the sentence extraction state and the partial output summary.

The proposed model consists of two parts, i.e., the document encoder and the sentence extractor.
The document encoder has a hierarchical architecture, which suits the compositionality of documents.
The sentence extractor is built with recurrent neural networks (RNN), which provides two main functionalities.
On one hand, the RNN is used to remember the partial output summary by feeding the selected sentence into it.
On the other hand, it is used to provide a sentence extraction state that can be used to score sentences with their representations.
At each  step during extraction, the sentence extractor reads the representation of the last extracted sentence.
It then produces a new sentence extraction state and uses it to score the relative importance of the rest sentences.

We conduct experiments on the \cnndm{} dataset.
The experimental results demonstrate that the proposed \ourModelName{} by jointly scoring and selecting sentences achieves significant improvements over  separated methods.
Our contributions are as follows:
\begin{itemize}
	\item We propose a joint sentence scoring and selection model for extractive document summarization.
	\item The proposed model can be end-to-end trained without handcrafted features.
	\item The proposed model significantly outperforms state-of-the-art methods  and achieves the best result on \cnndm{} dataset.
\end{itemize}

%% file: body/related.tex
\section{Related Work}
Extractive document summarization has been extensively studied for years.
As an effective approach, extractive methods are popular and dominate the summarization research.
Traditional extractive summarization systems use two key techniques to form the summary, sentence scoring and sentence selection.
Sentence scoring is critical since it is used to measure the saliency of a sentence.
Sentence selection is based on the scores of sentences to determine which sentence should be extracted, which is usually done heuristically.

Many techniques have been proposed to model and score sentences.
Unsupervised methods do not require model training or data annotation.
In these methods, many surface features are useful, such as term frequency \cite{luhn1958automatic}, TF*IDF weights \cite{erkan2004lexrank}, sentence length \cite{cao2015ranking} and sentence positions \cite{Ren:2017:LCS:3077136.3080792}.
These features can be used alone or combined with weights.

Graph-based methods \cite{erkan2004lexrank,mihalcea2004textrank,wan2006improved} are also applied broadly to ranking sentences.
In these methods, the input document is represented as a connected graph.
The vertices represent the sentences, and the edges between vertices have attached weights that show the similarity of the two sentences.
The score of a sentence is the importance of its corresponding vertex, which can be computed using graph algorithms.

Machine learning techniques are also widely used for better sentence modeling and importance estimation.
\citet{kupiec1995trainable} use a Naive Bayes classifier to learn feature combinations.
\citet{conroy2001text} further use a Hidden Markov Model in document summarization.
\citet{gillick2009scalable} find that using bigram features consistently yields better performance than unigrams or trigrams
for ROUGE \cite{lin2004rouge} measures.

\citet{carbonell1998use} proposed the Maximal Marginal Relevance (MMR) method as a heuristic in sentence selection.
Systems using MMR select the sentence which has the maximal score and is minimally redundant with previous selected sentences.
\citet{mcdonald2007study} treats sentence selection as an optimization problem under some constraints such as summary length.
Therefore, he uses Integer Linear Programming (ILP) to solve this optimization problem.
Sentence selection can also be seen as finding the optimal subset of sentences in a document.
\citet{lin2011class} propose using submodular functions to find the subset.

Recently, deep neural networks based approaches have become popular for extractive document summarization.
\citet{cao2015learning} develop a novel summary system called PriorSum, which applies  enhanced convolutional neural networks to capture the summary prior features derived from length-variable phrases.
\citet{Ren:2017:LCS:3077136.3080792} use a two-level attention mechanism to measure the contextual relations of sentences.
\citet{cheng-lapata:2016:P16-1} propose  treating document summarization as a sequence labeling task.
They first encode the sentences in the document and then classify each sentence into two classes, i.e., extraction or not.
\citet{nallapati2017summarunner} propose a system called SummaRuNNer with more features, which also treat extractive document summarization as a sequence labeling task.
The two works are both in the separated paradigm, as they first assign a probability of being extracted to each sentence, and then select sentences according to the probability  until reaching the length limit. 
\citet{ren2016redundancy} train two neural networks with handcrafted features. One is used to rank the sentences to select the first sentence, and the other one is used to model the redundancy during sentence selection.
However, their model of measuring the redundancy only considers the redundancy between the sentence that has the maximal score, which lacks the modeling of all the selection history.

%% file: body/problem.tex
\section{Problem Formulation}
\label{sec:problem}

Extractive document summarization aims to extract informative sentences to represent the important meanings of a document.
Given a document  $ \mathcal{D} = (S_{1}, S_{2}, \dots, S_{L}) $ containing $ L $ sentences, an extractive summarization system should select a subset of $ \mathcal{D} $ to form the output summary $ \mathcal{S} = \lbrace \hat{S}_{i} \vert \hat{S}_{i} \in \mathcal{D} \rbrace  $.
During the training phase, the reference summary $ \mathcal{S}^{*} $ and the  score of an output summary $ \mathcal{S} $ under a given evaluation function $ r(\mathcal{S} \vert \mathcal{S}^{*}) $ are available.
The goal of training is to learn a scoring function $ f(\mathcal{S}) $ which can be used to find the best summary during testing:

\begin{equation*}
\begin{aligned}
& \argmax_{\mathcal{S}}
& & f(\mathcal{S}) \\
& \text{s.t.}
&  \mathcal{S} &= \lbrace \hat{S}_{i} \vert \hat{S}_{i} \in \mathcal{D} \rbrace \\
&& \vert \mathcal{S} \vert &\leq l.
\end{aligned}
\end{equation*}
where $ l $ is length limit of the output summary.
In this paper, $ l $ is the sentence number limit.

Previous state-of-the-art summarization systems search the best solution using the learned scoring function $ f(\cdot) $ with two methods, MMR and ILP.
In this paper, we adopt the MMR method.
Since MMR tries to maximize the relative gain given previous extracted sentences, we let the model to learn to score this gain.
Previous works adopt \textsc{Rouge} recall as the evaluation $ r(\cdot) $ considering the DUC tasks have byte length limit for summaries.
In this work, we adopt the \cnndm{} dataset to train the neural network model, which does not have this length limit.
To prevent the tendency of choosing longer sentences, we use \textsc{Rouge} F1 as the evaluation function $ r(\cdot) $, and set the length limit $ l $ as a fixed number of sentences.

Therefore, the proposed model is trained to learn a scoring function $ g(\cdot) $ of the \textsc{Rouge} F1 gain, specifically:
\begin{empheq}{align}
g(S_{t} \vert \mathbb{S}_{t-1}) = r\left(\mathbb{S}_{t-1} \cup \lbrace S_{t} \rbrace\right) - r(\mathbb{S}_{t-1})
\end{empheq}
where $ \mathbb{S}_{t-1} $ is the set of previously selected sentences, and we omit the condition $ \mathcal{S}^{*} $ of $ r(\cdot) $ for simplicity.
At each time $ t $, the summarization system chooses the sentence with maximal \textsc{Rouge} F1 gain until reaching the sentence number limit.

%% file: body/model.tex
\section{Neural Document Summarization}

\begin{figure*}[htbp]
	\centering
	\includegraphics[scale=1]{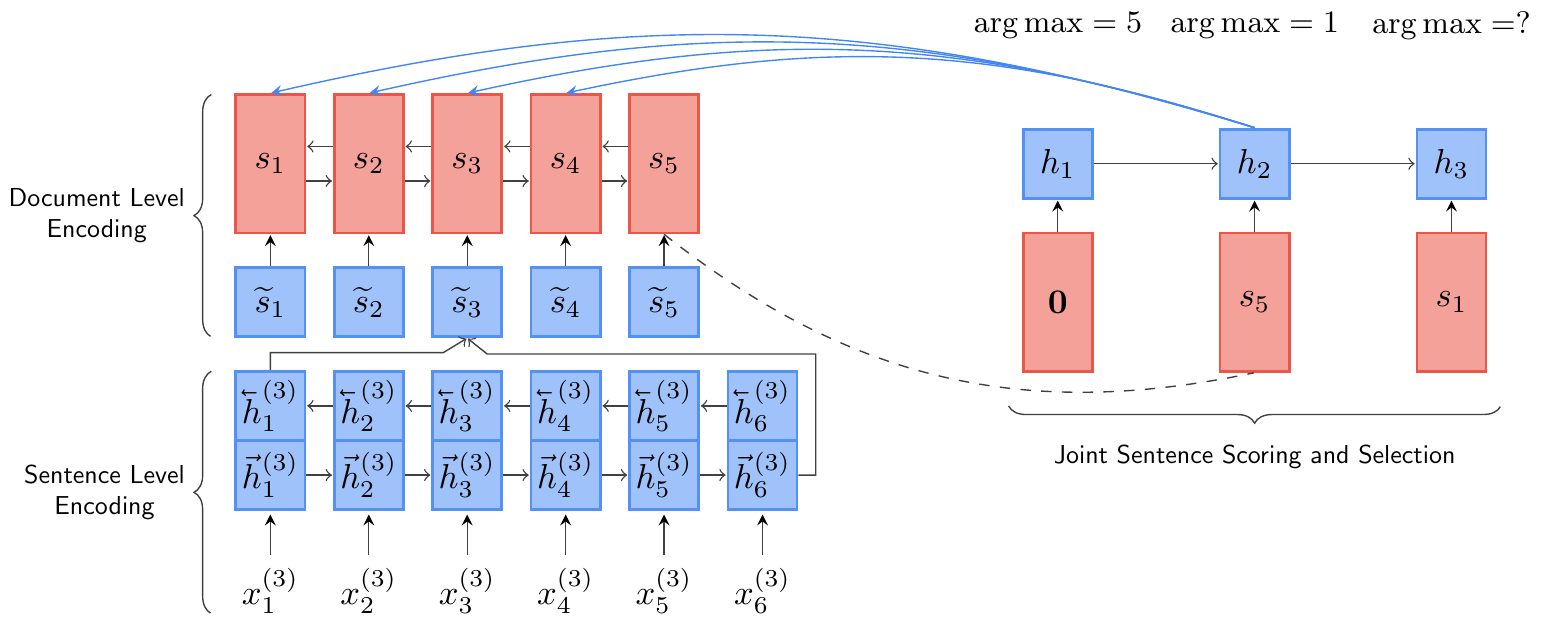}
	\caption{\label{fig:model} Overview of the \ourModelName{} model. The model extracts  $ S_{5} $ and $ S_{1} $ at the first two steps.
		At the first step, we feed the model a zero vector $ \mathbf{0} $ to represent empty partial output summary. At the second and third steps, the representations of previously selected sentences $ S_{5} $ and $ S_{1} $, i.e., $ s_{5} $ and $ s_{1} $, are fed into the extractor RNN. At the second step, the model only scores the first 4 sentences since the 5th one is already included in the partial output summary.}
\end{figure*}

Figure \ref{fig:model} gives the overview of \ourModelName{}, which consists of a hierarchical document encoder, and a sentence extractor.
Considering the intrinsic hierarchy nature of documents, that words form a sentence and sentences form a document, we employ a hierarchical document encoder to reflect this hierarchy structure.
The sentence extractor scores the encoded sentences and extracts one of them at each step until reaching the output sentence number limit.
In this section, we will first introduce the hierarchical document encoder, and then describe how the model produces summary by joint sentence scoring and selection.

\subsection{Document Encoding}
We employ a hierarchical document encoder to represent the sentences in the input document.
We encode the document in two levels, i.e., sentence level encoding and document level encoding.
Given a  document $ \mathcal{D} = (S_{1}, S_{2}, \dots, S_{L}) $ containing $ L $ sentences.
The sentence level encoder reads the $j$-th input sentence $ S_{j} = ( x_{1}^{(j)}, x_{2}^{(j)}, \dots ,x_{n_{j}}^{(j)}  ) $ and constructs the basic sentence representation $ \widetilde{s}_{j} $.
Here we employ a bidirectional GRU (BiGRU) \cite{cho-EtAl:2014:EMNLP2014} as the recurrent unit, where GRU is defined as:
\begin{empheq}{align}
z_i &= \sigma(\textbf{W}_z[x_i,h_{i-1}]) \\
r_i &= \sigma(\textbf{W}_r[x_i,h_{i-1}]) \\
\widetilde{h}_i &= \tanh(\textbf{W}_h[x_i,r_i \odot h_{i-1}]) \\
h_i &= (1-z_i)\odot h_{i-1} + z_i \odot \widetilde{h}_i 
\end{empheq}
where $ \textbf{W}_z $, $ \textbf{W}_r $ and $ \textbf{W}_h $ are weight matrices.

The BiGRU consists of a forward GRU and a backward GRU.
The forward GRU reads the word embeddings in  sentence $ S_{j} $ from left to right and gets a sequence of hidden states, $ (\vec{h}_{1}^{(j)}, \vec{h}_{2}^{(j)}, \dots, \vec{h}_{n_{j}}^{(j)})  $.
The backward GRU reads the input sentence embeddings reversely, from right to left, and results in another sequence of hidden states, $ (\cev{h}_{1}^{(j)}, \cev{h}_{2}^{(j)}, \dots, \cev{h}_{n_{j}}^{(j)}) $:
\begin{empheq}{align}
\vec{h}_{i}^{(j)} &=\text{GRU}(x_{i}^{(j)}, \vec{h}_{i-1}^{(j)})\\
\cev{h}_{i}^{(j)} &= \text{GRU}(x_{i}^{(j)}, \cev{h}_{i+1}^{(j)})
\end{empheq}
where the initial states of the BiGRU are set to zero vectors, i.e., $ \vec{h}_{1}^{(j)} = 0 $ and $ \cev{h}_{n_{j}}^{(j)} = 0 $.

After reading the words of the sentence $ S_{j} $, we construct its sentence level  representation $ \widetilde{s}_{j} $ by concatenating the last forward and backward GRU hidden vectors:
\begin{equation}
\widetilde{s}_{j} = \left[
\begin{matrix}
\cev{h}_{1}^{(j)}\\
\vec{h}_{n_{j}}^{(j)}
\end{matrix}
\right]
\end{equation}

We use another BiGRU as the document level encoder to read the sentences.
With the sentence level encoded vectors $  (\widetilde{s}_{1}, \widetilde{s}_{2}, \dots, \widetilde{s}_{L}) $ as inputs, the document level encoder does forward and backward GRU encoding and produces two list of hidden vectors: $ (\vec{s}_{1}, \vec{s}_{2}, \dots, \vec{s}_{L}) $ and $ (\cev{s}_{1}, \cev{s}_{2}, \dots, \cev{s}_{L}) $.
The document level representation $ s_{i} $ of sentence $ S_{i} $ is the concatenation of the forward and backward hidden vectors:
\begin{equation}
s_{i} = \left[
\begin{matrix}
\vec{s}_{i}\\
\cev{s}_{i}
\end{matrix}
\right]
\end{equation}
We then get the final sentence vectors in the given document: $ D = (s_{1}, s_{2}, \dots, s_{L}) $.
We use sentence $ S_{i} $ and its representative vector $ s_{i} $ interchangeably in this paper.

\subsection{Joint Sentence Scoring and Selection}
Since the separated sentence scoring and selection cannot utilize the information of each other, the goal of our model is to make them benefit each other.
We couple these two steps together so that: a) sentence scoring can be aware of previously selected sentences; b) sentence selection can be simplified since the scoring function is learned to be the \textsc{Rouge} score gain as described in section \ref{sec:problem}.

Given the last extracted sentence $ \hat{S}_{t-1} $, the sentence extractor decides the next sentence $ \hat{S}_{t} $  by scoring the remaining document sentences.
To score the document sentences considering both their importance and partial output summary, the model should have two key abilities: 1) remembering the information of previous selected sentences; 2) scoring the remaining document sentences based on both the previously selected sentences and the importance of remaining sentences.
Therefore, we employ another GRU as the recurrent unit to remember the partial output summary, and use a Multi-Layer Perceptron (MLP) to score the document sentences.
Specifically, the GRU takes the document level representation $ s_{t-1} $ of the last  extracted sentence $ \hat{S}_{t-1} $ as input to produce its current hidden state $ h_{t} $.
The sentence scorer, which is a two-layer MLP, takes two input vectors, namely the current hidden state $ h_{t} $ and the sentence representation vector $ s_{i} $, to calculate the score $ \delta(S_{i}) $ of sentence $ S_{i} $.
\begin{empheq}{align}
h_{t} &= \text{GRU}(s_{t-1}, h_{t-1} ) \\
\delta (S_{i}) &= \mathbf{W}_{s} \tanh\left(  \mathbf{W}_{q}h_{t} + \mathbf{W}_{d}s_{i} \right)
\end{empheq}
where $ \mathbf{W}_{s} $, $ \mathbf{W}_{q} $ and $ \mathbf{W}_{d} $ are learnable parameters, and we omit the bias parameters for simplicity.

When extracting the first sentence, we initialize the GRU hidden state $ h_{0} $ with a linear layer with tanh activation function:
\begin{empheq}{align}
h_{0} &= \tanh \left( \mathbf{W}_{m} \cev{s}_{1} + b_{m} \right)\\
S_{0} &= \varnothing  \\
s_{0} &= \mathbf{0}
\end{empheq}
where$ \mathbf{W}_{m} $ and $ b_{m} $ are learnable parameters, and $ \cev{s}_{1} $ is the last backward state of the document level encoder BiGRU.
Since we do not have any sentences extracted yet, we use a zero vector to represent the previous extracted sentence, i.e., $ s_{0} = \mathbf{0} $.

With the scores of all sentences at time $ t $, we choose the sentence with maximal gain score:
\begin{empheq}{align}
\hat{S}_{t} &= \argmax_{S_{i} \in \mathcal{D}} \delta (S_{i})
\end{empheq}

\subsection{Objective Function}
Inspired by \citet{inan2016tying}, we optimize the Kullback-Leibler (KL) divergence of the model prediction $ P $ and the labeled training data distribution $ Q $.
We normalize the predicted sentence score $ \delta(S_{i}) $ with softmax function to get the model prediction distribution $ P $:
\begin{eqnarray}
P( \hat{S}_{t} = S_{i} ) = \frac{\exp\left(\delta(S_{i})\right)}{\sum_{k=1}^{L}\exp\left(\delta(S_{k})\right)}
\end{eqnarray}

During training, the model is expected to learn the relative \textsc{Rouge} F1 gain at time step $ t $ with previously selected sentences $ \mathbb{S}_{t-1} $.
Considering that the F1 gain value might be negative in the labeled data, we follow previous works \citep{Ren:2017:LCS:3077136.3080792} to use Min-Max Normalization to rescale the gain value to $ [0, 1] $:
\begin{empheq}{align}
g(S_{i}) &= r(  \mathbb{S}_{t-1} \cup  \lbrace S_{i} \rbrace) -  r(\mathbb{S}_{t-1}) \\
\widetilde{g}(S_{i}) &= \frac{g(S_{i}) - \min\left(g(S)\right)}{\max\left(g(S)\right)- \min\left(g(S)\right)}
\end{empheq}
We then apply a softmax operation with temperature $ \tau $ \citep{hinton2015distilling} \footnote{We set $ \tau = 20 $ empirically according to the model performance on the development set. } to produce the labeled data distribution $ Q $ as the training target.
We apply the temperature $ \tau $ as a smoothing factor to produce a smoothed label distribution $ Q $: 
\begin{empheq}{align}
\label{eq:tau_eq}
	Q(S_{i}) = \frac{\exp\left( \tau \widetilde{g}({S_{i}})\right)}{\sum_{k=1}^{L}\exp\left( \tau \widetilde{g}({S_{k}})\right)}
\end{empheq}

Therefore, we minimize the \textit{KL} loss function $ J $:
\begin{eqnarray}
\label{eq:loss}
J = \dKL{P}{Q}
\end{eqnarray}

%% file: body/exp.tex
\section{Experiments}

\subsection{Dataset}
\label{sec:dataset}
A large scale dataset is essential for training neural network-based summarization models.
We use the \cnndm{} dataset \citep{hermann2015teaching, nallapatiabstractive} as the training set in our experiments.
The \cnndm{} news contain articles and their corresponding highlights.
The highlights are created by human editors and are abstractive summaries.
Therefore, the highlights are not ready for training extractive systems due to the lack of supervisions.

We create an extractive summarization training set based on \textit{CNN/Daily Mail} corpus.
To determine the sentences to be extracted, we design a rule-based system to label the sentences in a given document similar to \citet{nallapati2017summarunner}.
Specifically, we construct training data by maximizing the \textsc{Rouge}-2 F1 score.
Since it is computationally expensive to find the global optimal combination of sentences, we employ a greedy approach.
Given a document with $ n $ sentences, we enumerate the candidates from $ 1 $-combination $ \binom{n}{1} $ to $ n $-combination $ \binom{n}{n} $.
We stop searching if the highest \textsc{Rouge}-2 F1 score in $ \binom{n}{k} $ is less than the best one in $ \binom{n}{k-1} $.
Table \ref*{table:data_stat} shows the data statistics of the \cnndm{} dataset.

We conduct data preprocessing using the same method\footnote{https://github.com/abisee/cnn-dailymail} in \citet{see-liu-manning:2017:Long}, including sentence splitting and word tokenization.
Both \citet{nallapatiabstractive,nallapati2017summarunner} use the \textit{anonymized} version of the data, where the named entities are replaced by identifiers such as \texttt{entity4}.
Following \citet{see-liu-manning:2017:Long}, we use the \textit{non-anonymized} version so we can directly operate on the original text.

\begin{table}[htbp]
	\begin{center}
		\begin{tabular}{@{~}l@{\hspace{1ex}}c@{\hspace{1ex}}c@{\hspace{1ex}}c@{~}}
			\toprule
			\bf \cnndm{}       & \bf Training & \bf Dev & \bf Test \\ 
			\midrule
			\#(Document)       & 287,227      & 13,368  &  11,490 \\
			\#(Ref / Document) & 1            & 1       & 1 \\
			Doc Len (Sentence) & 31.58        & 26.72   & 27.05 \\
			Doc Len (Word)     & 791.36       & 769.26  & 778.24 \\
			Ref Len (Sentence) & 3.79         & 4.11    & 3.88   \\
			Ref Len (Word)     & 55.17        & 61.43   & 58.31 \\
			\bottomrule
		\end{tabular}
	\end{center}
	\caption{\label{table:data_stat}Data statistics of \cnndm{} dataset.}
\end{table}

\subsection{Implementation Details}

\paragraph{Model Parameters}
The vocabulary is collected from the \cnndm{} training data.
We lower-case the text and there are 732,304 unique word types.
We use the top 100,000 words as the model vocabulary since they can cover 98.23\% of the training data.
The size of word embedding, sentence level encoder GRU, document level encoder GRU are set to 50, 256, and 256 respectively.
We set the sentence extractor GRU hidden size to 256.

\paragraph{Model Training}
We initialize the model parameters randomly using a Gaussian distribution with Xavier scheme \citep{glorot2010understanding}.
The word embedding matrix is initialized using pre-trained 50-dimension GloVe vectors \citep{pennington2014glove}\footnote{\url{https://nlp.stanford.edu/projects/glove/}}.
We found that larger size GloVe does not lead to improvement. Therefore, we use 50-dim word embeddings for fast training. 
The pre-trained GloVe vectors contain 400,000 words and cover 90.39\% of our model vocabulary.
We initialize the rest of the word embeddings randomly using a Gaussian distribution with Xavier scheme.
The word embedding matrix is not updated during training.
We use Adam \citep{kingma2014adam} as our optimizing algorithm.
For the hyperparameters of Adam optimizer, we set the learning rate $ \alpha = 0.001 $, two momentum parameters $ \beta_{1} = 0.9 $ and $ \beta_{2} = 0.999 $ respectively, and $ \epsilon=10^{-8} $.
We also apply gradient clipping \citep{pascanu2013difficulty} with range $ [-5, 5] $ during training.
We use dropout \citep{srivastava2014dropout} as regularization with probability $ p = 0.3 $ after the sentence level encoder and $ p=0.2 $ after the document level encoder.
We truncate each article to 80 sentences and each sentence to 100 words during both training and testing.
The model is implemented with PyTorch \citep{paszke2017automatic}.
We release the source code and related resources at \url{https://res.qyzhou.me}.

\paragraph{Model Testing}
At test time, considering that \textsc{LEAD3} is a commonly used and strong extractive baseline,  we make \ourModelName{} and the baselines extract 3 sentences to make them all comparable.

\subsection{Baseline}

We compare \ourModelName{} model with the following state-of-the-art baselines:
\begin{description}
	\item[\textsc{LEAD3}] The commonly used baseline by selecting the first three sentences as the summary.
	\item[\textsc{TextRank}]  An unsupervised algorithm based on weighted-graphs proposed by \citet{mihalcea2004textrank}. We use the implementation in Gensim \citep{rehurek_lrec}.
	\item[\textsc{CRSum}] \citet{Ren:2017:LCS:3077136.3080792} propose an extractive summarization system which considers the contextual information of a sentence.
	 We train this baseline model with the same training data as our approach.
	\item[\textsc{NN-SE}] \citet{cheng-lapata:2016:P16-1} propose an extractive system which models document summarization as a sequence labeling task. We train this baseline model with the same training data as our approach.
	\item[\textsc{SummaRuNNer}] \citet{nallapati2017summarunner} propose to add some interpretable features such as sentence absolute and relative positions.
	\item[\textsc{PGN}] Pointer-Generator Network (PGN). A state-of-the-art abstractive document summarization system proposed by \citet{see-liu-manning:2017:Long}, which incorporates copying and coverage mechanisms.
\end{description}

\subsection{Evaluation Metric}
We employ \textsc{Rouge} \citep{lin2004rouge} as our evaluation metric.
\textsc{Rouge} measures the quality of summary by computing overlapping lexical units, such as unigram, bigram, trigram, and longest common subsequence (LCS).
It has become the standard evaluation metric for DUC shared tasks and popular for summarization evaluation.
Following previous work, we use \textsc{Rouge}-1 (unigram), \textsc{Rouge}-2 (bigram) and \textsc{Rouge}-L (LCS) as the evaluation metrics in the reported experimental results.

\subsection{Results}

We use the official ROUGE script\footnote{\url{http://www.berouge.com/}} (version 1.5.5) to evaluate the summarization output.
Table \ref{table:cnndm_result} summarizes the results on \cnndm{} data set using full length \textsc{Rouge}-F1\footnote{The ROUGE evaluation option is, -m -n 2} evaluation.
It includes two unsupervised baselines, \textsc{LEAD3} and \textsc{TextRank}.
The table also includes three state-of-the-art neural network based extractive models, i.e., \textsc{CRSum}, \textsc{NN-SE} and \textsc{SummaRuNNer}.
In addition, we report the state-of-the-art abstractive \textsc{PGN} model.
The result of \textsc{SummaRuNNer} is on the \textit{anonymized} dataset and not strictly comparable to our results on the \textit{non-anonymized} version dataset.
Therefore, we also include the result of \textsc{LEAD3} on the \textit{anonymized} dataset as a reference.

\begin{table}[h]
	\begin{center}
		\begin{tabular}{@{~}l@{\hspace{1ex}}c@{\hspace{1ex}}c@{\hspace{1ex}}c@{~}}
			\toprule
			\bf Models & \bf {\small \textsc{Rouge-1}} & {\small \bf \textsc{Rouge-2}} & {\small \bf \textsc{Rouge-L}} \\ 
			\midrule
			\textsc{LEAD3}  & 40.24\significant{}  & 17.70\significant{} & 36.45\significant{}  \\
			\textsc{TextRank} &  40.20\significant{} & 17.56\significant{}  & 36.44\significant{} \\
			\textsc{CRSum}  & 40.52\significant{} & 18.08\significant{}  & 36.81\significant{} \\
			\textsc{NN-SE}    & 41.13\significant{}  &  18.59\significant{}  & 37.40\significant{}   \\
			\textsc{PGN}\otherpaper{}  & 39.53\significant{}  &  17.28\significant{} &  36.38\significant{} \\
			\hline
			\small \textsc{LEAD3}\otherpaper{} * & 39.2  & 15.7 & 35.5  \\
			{\small \textsc{SummaRuNNer}\otherpaper{}} * & 39.6 &  16.2  & 35.3 \\
			\hline
			\textbf{\ourModelName{}}  & \textbf{41.59}  & \textbf{19.01}  &  \textbf{37.98} \\
			\bottomrule
		\end{tabular}
	\end{center}
	\caption{Full length \label{table:cnndm_result}\textsc{Rouge} F1  evaluation (\%) on \cnndm{} test set. Results with \otherpaper{} mark are taken from
		the corresponding papers. Those marked with * were trained and evaluated on the anonymized dataset, and so are not strictly comparable to our results on the original text. 
All our \textsc{Rouge} scores have a 95\% confidence interval of at most $ \pm $0.22 as reported by the official ROUGE script.
The improvement is statistically significant with respect to the results with superscript \significant{} mark.	 }
\end{table}

\ourModelName{} achieves 19.01 \textsc{Rouge}-2 F1 score on the \cnndm{} dataset.
Compared to the unsupervised baseline methods, \ourModelName{} performs better by a large margin.
In terms of \textsc{Rouge}-2 F1, \ourModelName{} outperforms the strong baseline \textsc{LEAD3} by 1.31 points.
\ourModelName{} also outperforms the neural network based models.
Compared to the state-of-the-art extractive model \textsc{NN-SE} \citep{cheng-lapata:2016:P16-1}, \ourModelName{} performs significantly better in terms of \textsc{Rouge}-1, \textsc{Rouge}-2 and \textsc{Rouge}-L F1 scores.
Shallow features, such as sentence position, have proven effective in document summarization \citep{Ren:2017:LCS:3077136.3080792,nallapati2017summarunner}.
Without any hand-crafted features, \ourModelName{} performs better than the \textsc{CRSum} and \textsc{SummaRuNNer} baseline models with features.
As given by the 95\% confidence interval in the official \textsc{Rouge} script,
our model achieves statistically significant improvements over all the baseline models.
To the best of our knowledge, the proposed \ourModelName{} model achieves the best results on the \cnndm{} dataset.

\begin{table}[h]
	\begin{center}
		\begin{tabular}{llll}
			\toprule
			\bf Models & Info & Rdnd & Overall \\ 
			\midrule
			\textsc{NN-SE}    & 1.36  &  1.29  & 1.39   \\
			\hline
			\textbf{\ourModelName{}}  & \textbf{1.33} & \textbf{1.21}  &  \textbf{1.34} \\
			\bottomrule
		\end{tabular}
	\end{center}
	\caption{\label{table:human}
	Rankings of \ourModelName{} and \textsc{NN-SE} in terms of informativeness (Info), redundancy (Rdnd) and overall quality by human participants  (lower is better).	 }
\end{table}

We also provide human evaluation results on a sample of test set.
We random sample 50 documents and ask three volunteers to evaluate the output of \ourModelName{} and the \textsc{NN-SE} baseline models.
They are asked to rank the output summaries from best to worst (with ties allowed) regarding  informativeness, redundancy and overall quality.
Table \ref{table:human} shows the human evaluation results.
\ourModelName{} performs better than the \textsc{NN-SE} baseline on all three aspects, especially in redundancy.
This indicates that by jointly scoring and selecting sentences, \ourModelName{} can produce summary with less content overlap since it re-estimates the saliency of remaining sentences considering both their contents and previously selected sentences.

%% file: body/diss.tex
\section{Discussion}

\begin{figure*}[htbp]
	\centering
	\includegraphics[width=\textwidth]{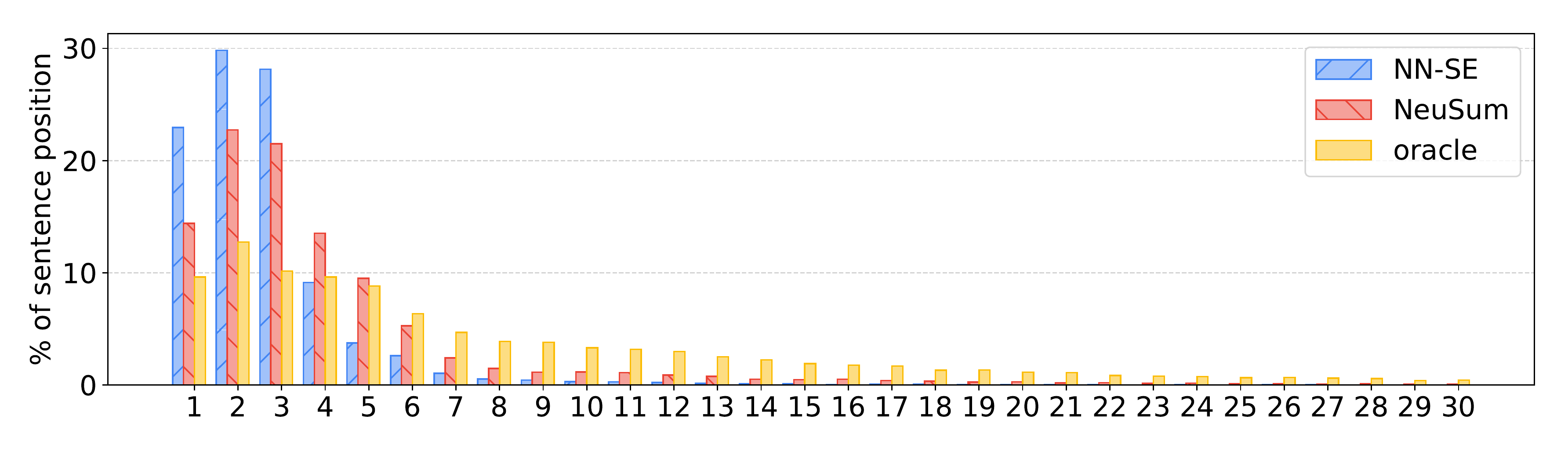}
	\caption{\label{fig:sent_position} Position distribution of selected sentences of the \textsc{NN-SE} baseline, our \ourModelName{} model and oracle on the test set. We only draw the first 30 sentences since the average document length is 27.05.}
\end{figure*}

\subsection{Precision at Step-$ t $}
\label{sec:prection_at_t}
We analyze the accuracy of sentence selection at each step.
Since we extract 3 sentences at test time, we show how \ourModelName{} performs when extracting each sentence.
Given a document $ D $ in test set $ \mathfrak{T} $, \ourModelName{} predicted summary $ \mathcal{S} $, its reference summary $ \mathcal{S}^{*} $, and the extractive oracle summary $ \mathcal{O} $ with respect to $ D $ and $ \mathcal{S}^{*} $ (we use the method described in section \ref{sec:dataset} to construct $ \mathcal{O} $), 
we define the precision at step $ t $ as $ p(\text{@}t) $:
\begin{equation}
	p(\text{@}t) = \frac{1}{\vert \mathfrak{T} \vert } \sum_{D \in \mathfrak{T}} \mathbf{1}_{\mathcal{O}}(\mathcal{S}[t])
\end{equation}
where $ \mathcal{S}[t] $ is the sentence extracted at step $ t $, and $ \mathbf{1}_{\mathcal{O}} $ is the indicator function defined as:
\begin{equation}
\mathbf{1}_{\mathcal{O}}(x) = \begin{cases}
1 &\text{if $x \in \mathcal{O}$}\\
0 &\text{if $x \notin \mathcal{O}$}
\end{cases}
\end{equation}

\begin{figure}[htbp]
	\centering
	\includegraphics[width=0.5\textwidth]{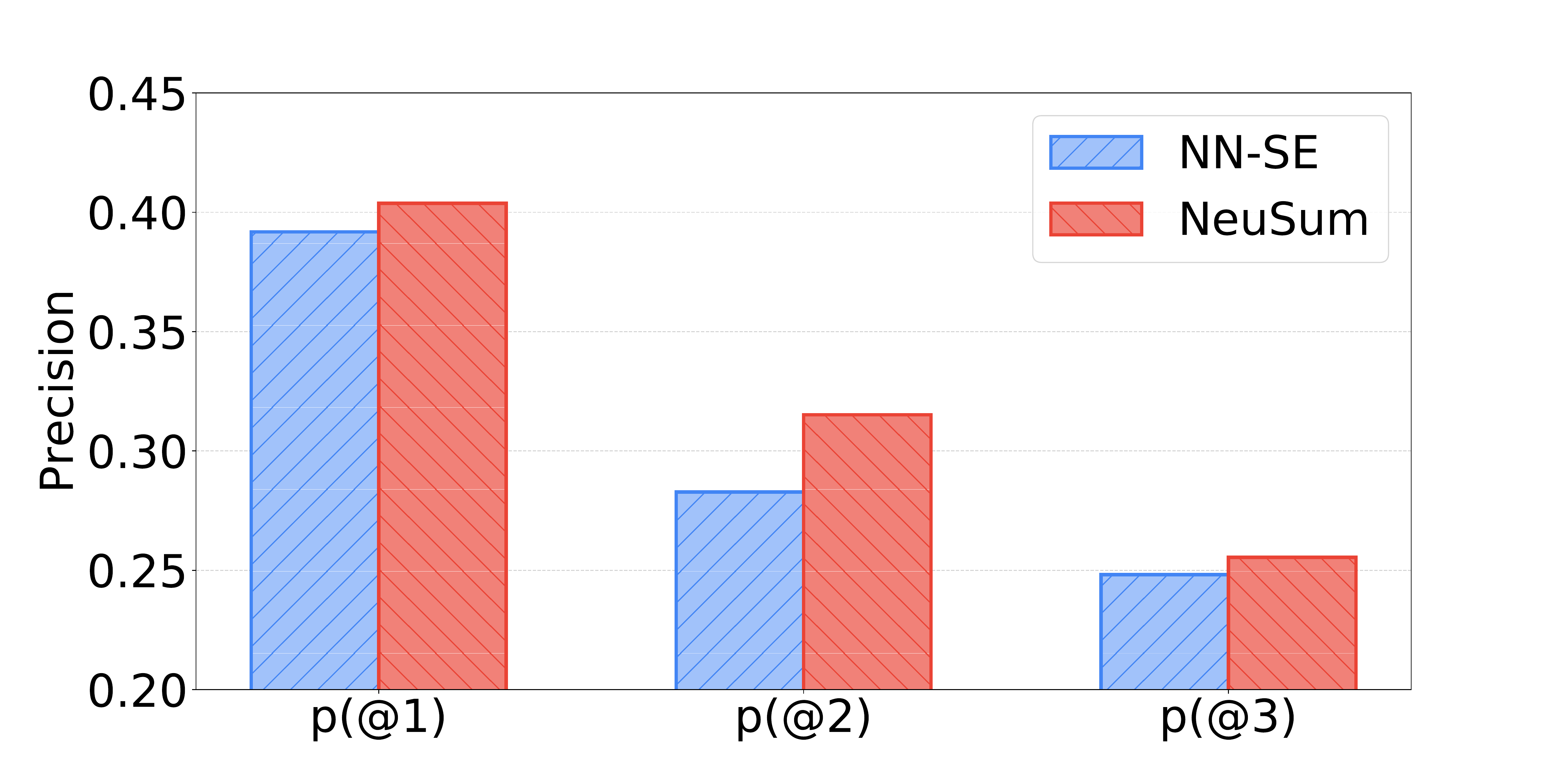}
	\caption{\label{fig:p_at_t} Precision of extracted sentence at step $ t $ of the \textsc{NN-SE} baseline and the \ourModelName{} model.}
\end{figure}

Figure \ref{fig:p_at_t} shows the precision at step $ t $  of \textsc{NN-SE} baseline and our \ourModelName{}.
It can be observed that \ourModelName{} achieves better precision than the \textsc{NN-SE} baseline at each step.
For the first sentence, both \ourModelName{} and \textsc{NN-SE} achieves good performance.
The \textsc{NN-SE} baseline has 39.18\% precision at the first step, and \ourModelName{} outperforms it by 1.2 points.
At the second step, \ourModelName{} outperforms \textsc{NN-SE} by a large margin.
In this step, the \ourModelName{} model extracts 31.52\% sentences correctly, which is 3.24 percent higher than 28.28\% of \textsc{NN-SE}.
We think the second step selection benefits from the first step in \ourModelName{} since it can remember the selection history, while the separated models lack this ability.

However, we can notice the trend that the precision drops fast after each selection.
We think this is due to two main reasons.
First, we think that the error propagation leads to worse selection for the third selection. As shown in Figure 2, the $ p(\text{@}1) $ and $ p(\text{@}2) $ are 40.38\% and 31.52\% respectively, so the history is less reliable for the third selection.
Second, intuitively, we think the later selections are more difficult compared to the previous ones since the most important sentences are already selected.

\subsection{Position of Selected Sentences}

Early works \cite{Ren:2017:LCS:3077136.3080792,nallapati2017summarunner} have shown that sentence position is an important feature in extractive document summarization. 
Figure \ref{fig:sent_position} shows the position distributions of the \textsc{NN-SE} baseline, our \ourModelName{} model and oracle on the \cnndm{} test set.
It can be seen that the \textsc{NN-SE} baseline model tends to extract large amount of leading sentences, especially the leading three sentences.
According to the statistics, about 80.91\% sentences selected by \textsc{NN-SE} baseline are in leading three sentences.

In the meanwhile, our \ourModelName{} model selects 58.64\% leading three sentences.
We can notice that in the oracle, the percentage of selecting leading sentences (sentence 1 to 5) is moderate, which is around 10\%.
Compared to \textsc{NN-SE}, the position of selected sentences in \ourModelName{} is closer to the oracle.
Although \ourModelName{} also extracts more leading sentences than the oracle, it selects more tailing ones.
For example, our \ourModelName{} model extracts more than 30\% of sentences in the range of sentence 4 to 6.
In the range of sentence 7 to 13, \textsc{NN-SE} barely extracts any sentences, but our \ourModelName{} model still extract sentences in this range.
Therefore, we think this is one of the reasons why \ourModelName{} performs better than \textsc{NN-SE}.

We analyze the sentence position distribution and offer an explanation for these observations.
Intuitively, leading sentences are important for a well-organized article, especially for newswire articles.
It is also well known that \textsc{LEAD3} is a very strong baseline.
In the training data, we found that 50.98\% sentences labeled as ``should be extracted'' belongs to the first 5 sentences, which may cause the trained model tends to select more leading sentences.
One possible situation is that one sentence in the tail of a document is more important than the leading sentences, but the margin between them is not large enough.
The models which separately score and select sentences might not select sentences in the tail whose scores are not higher than the leading ones.
These methods may choose the safer leading sentences as a fallback in such confusing situation because there is no direct competition between the leading and tailing candidates.
In our \ourModelName{} model, the scoring and selection are jointly learned, and at each step the tailing candidates can compete directly with the leading ones.
Therefore, \ourModelName{} can be more discriminating when dealing with this situation.

%% file: body/conclusion.tex
\section{Conclusion}
Conventional approaches to extractive document summarization contain two separated steps: sentence scoring and sentence selection.
In this paper, we present a novel neural network framework for extractive document summarization by jointly learning to score and select sentences to address this issue.
The most distinguishing feature of our approach from previous methods is that it combines sentence scoring and selection into one phase.
Every time it selects a sentence, it scores the sentences according to the partial output summary and current extraction state.
\textsc{Rouge} evaluation results show that the proposed joint sentence scoring and selection approach significantly outperforms previous separated methods.

%% file: body/ack.tex
\section*{Acknowledgments}

We thank three anonymous reviewers for their helpful comments.
We also thank  Danqing Huang,  Chuanqi Tan, Zhirui Zhang, Shuangzhi Wu and Wei Jia  for helpful discussions.
The work of this paper is funded by the project of National Key Research and Development Program of China (No. 2017YFB1002102) and the project of National Natural Science Foundation of China (No. 91520204). 
The first author is funded by the Harbin Institute of Technology Scholarship Fund.